\documentclass[11pt,a4paper]{article}
\usepackage[utf8]{inputenc}
\usepackage[T1]{fontenc}
\usepackage{times}
\usepackage[margin=1in]{geometry}
\usepackage{amsmath,amssymb}
\usepackage{booktabs}
\usepackage{graphicx}
\usepackage{hyperref}
\usepackage{natbib}
\usepackage{xcolor}
\usepackage{tabularx}
\usepackage{multirow}

\hypersetup{colorlinks=true,linkcolor=blue,citecolor=blue,urlcolor=blue}

\title{Before You Interpret the Profile:\\Validity Scaling for LLM Metacognitive Self-Report}

\author{Jon-Paul Cacioli\\
Independent Researcher, Melbourne, Australia\\
ORCID: 0009-0000-7054-2014\\
\texttt{https://github.com/synthiumjp/validity-scaling-llm}}

\date{}

\begin{document}
\maketitle

\begin{abstract}
Clinical personality assessment screens response validity before interpreting substantive scales. LLM evaluation does not. We apply the validity scaling framework from the PAI \citep{morey1991,morey2007} and MMPI-3 \citep{benporath2020} to metacognitive probe data from 20 frontier models across 524 items. Six validity indices are operationalised: L (maintaining confidence on errors), K (betting on errors), F (withdrawing consensus-endorsed items), Fp (withdrawing correct answers), RBS (inverted monitoring), and TRIN (fixed responding). These are calibrated against a 20-model derivation sample with family-level comparison groups. A tiered classification system, validated against eight synthetic response policies, identifies four models as construct-level invalid and two as elevated.

The indices show strong convergent validity (L--K: $r = .82$; F--Fp: $r = .99$; dual-probe convergence: $r = .88$) and appropriate discriminant structure (L--F: $r = -.10$). Valid-profile models produce item-sensitive confidence, with KEEP decisions tracking correctness (mean $r = .18$, 14 of 16 significant). Invalid-profile models do not (mean $r = -.20$). The group difference is large ($d = 2.17$, $p = .001$) and robust to leave-one-out removal of any single model. Chain-of-thought training produces two opposite response distortions: systematic over-withdrawal (DeepSeek-R1: 37\% WITHDRAW+BET contradiction on correct items) and blanket confidence (Qwen Think: $L = .974$). Retrospective and prospective probes yield inconsistent validity profiles in 12 of 20 models. Two latent dimensions, corresponding to the under-reporting and over-reporting blocs of clinical assessment, account for 94.6\% of index variance.

We argue the mapping from clinical validity to LLM evaluation is structural, not anthropomorphic. The methodology screens response patterns for interpretability regardless of what produces them. The operational principle is simple: screen before you interpret.
\end{abstract}

\section{Introduction}

\subsection{The problem}

Every LLM metacognition benchmark treats the model's confidence signal as informative. Downstream systems use these signals for abstention, routing, and human-in-the-loop decisions. Nobody checks whether the signal actually tracks correctness before building on it.

Clinical assessment solved this problem decades ago. The PAI \citep{morey1991,morey2007} and MMPI-3 \citep{benporath2020} include validity scales that detect distorted response patterns before substantive scales are interpreted. In forensic, clinical, and personnel selection contexts, a profile with elevated validity scales is flagged as uninterpretable regardless of what the clinical scales show. This is not optional. It is the first step of any competent interpretation.

The three metacognitive profiles identified in \citet{cacioli2026c} across 20 frontier LLMs represent, in clinical terms, three response styles. Two of them produce confidence signals that contain no item-level information. The present paper provides the formal methodology for detecting this.

The central question is whether a model's confidence signal carries item-level information about correctness. Validity indices are the diagnostic machinery for answering that question. This is a methods paper. We introduce a validity-screening framework that can be applied to any metacognitive probe dataset.

\subsection{The structural parallel}

We are not claiming that LLMs have personalities, experience doubt, or engage in impression management. The mapping from clinical validity to LLM evaluation is structural, not psychological.

What is shared between clinical self-report and LLM probe data is the data format (binary responses to structured items), the problem (response patterns dominated by a general strategy rather than item-specific information), the consequence (if the pattern is dominated by a response set, the data is uninformative), and the solution (screen for validity before interpreting content).

What is not shared is the cause of response distortion. In humans, distortion arises from motivation, psychopathology, or deliberate malingering. In LLMs, it arises from training objectives, RLHF, or architectural constraints. We make no claims about LLM phenomenology. We make no claims about LLM experience of doubt or confidence. The methodology does not require these claims. It requires only that the data be structured as self-report and that the question ``does this signal carry information?'' be answerable.

The validity framework is a statistical methodology for detecting whether response patterns carry item-level information. A random number generator that produced 95\% KEEP would be flagged by the same indices, not because it has a personality disorder but because its data is uninformative. The methodology is substrate-independent.

This parallels the application of signal detection theory to LLMs \citep{cacioli2026a}. SDT was developed for human psychophysics, but $d'$ measures sensitivity regardless of what the detector is. The methodology transfers because the measurement problem transfers.

\subsection{The policy objection}

A sceptical reader will argue that LLM KEEP/WITHDRAW/BET decisions are policy outputs, not introspective reports. They are trained decision rules. Therefore ``response validity'' is the wrong construct, and what we observe is simply different policies.

We agree that LLMs produce policy outputs. The validity framework does not require introspection. It requires only one thing: that the confidence signal carries item-level information about correctness. A policy that produces KEEP on 99\% of items regardless of correctness is a policy whose output is uninformative for any downstream use that depends on confidence discriminating correct from incorrect responses.

To be explicit about what this paper does not claim: we do not claim LLMs introspect, we do not claim LLMs have personalities or psychological states, we do not claim clinical constructs transfer psychologically, and we do not claim validity indices measure latent internal states. We claim only that some confidence signals carry item-level information and others do not, and that this distinction can be detected.

We retain the term ``validity'' because the statistical structure of the problem is identical to the structure addressed by clinical validity scales: detecting when response patterns suppress item-level information.

\subsection{Prior work}

Several lines of work address LLM confidence and metacognition. \citet{xiong2024} surveyed confidence elicitation methods and found pervasive overconfidence when LLMs verbalise confidence, predominantly in the 80--100\% range. \citet{kadavath2022} showed that LLMs can sometimes discriminate questions they answer correctly from those they do not. \citet{steyvers2025} reviewed LLM metacognition through AUROC and meta-$d'$. \citet{ackerman2025} tested metacognitive control using a second-chance paradigm and found limited strategic deployment. \citet{dai2026} demonstrated that verbal confidence scales suffer from severe discretisation. \citet{kumaran2025} identified a pronounced choice-supportive bias that reinforces initial confidence. \citet{scholten2024} argued that LLMs exhibit ``metacognitive myopia,'' lacking monitoring and control processes that produce systematic biases. \citet{phillips2026} introduced a decision-theoretic metric for LLM confidence reliability, showing that even frontier models remain prone to overconfidence on complex tasks. \citet{cacioli2026b} applied Type-2 SDT to decompose metacognitive efficiency from task performance, finding fully inverted rankings. \citet{cacioli2026c} identified three behavioural profiles across 20 frontier LLMs using a cross-domain monitoring battery.

An emerging field of ``LLM Psychometrics'' \citep{lin2025} applies psychometric instruments and principles to LLM evaluation broadly, including personality, values, and cognitive biases. Our work contributes to this field by importing the specific methodology of clinical validity scaling, which has not previously been operationalised for LLM confidence data.

None of the metacognition studies above screens for response validity before analysing the confidence signal. They all assume the data is interpretable.

\subsection{The validity scaling framework in clinical assessment}

Clinical personality assessment uses validity scales organised into three categories. Content non-responsiveness scales detect inconsistent or fixed responding: VRIN and TRIN on the MMPI-3 \citep{benporath2020}, ICN and INF on the PAI \citep{morey1991,morey2007}. Over-reporting scales detect exaggeration or feigning: F, Fp, Fs, FBS, and RBS on the MMPI-3, NIM and MAL on the PAI. Under-reporting scales detect minimisation or positive impression management: L and K on the MMPI-3, PIM and DEF on the PAI.

These scales are interpreted before any substantive clinical scales. If validity is compromised, the substantive profile is flagged as uninterpretable \citep{benporath2020}. Thresholds are set using normative reference groups and comparison samples from specific populations (forensic, clinical, public safety). The Rogers Discriminant Function \citep{rogers2003} provides a trained multivariate discriminant for detecting feigned profiles. Coaching effects in forensic assessment \citep{sellbom2008} demonstrate that response preparation systematically elevates under-reporting scales, a structural parallel to RLHF training effects.

\subsection{Contributions}

We make seven contributions. First, six validity indices for LLM metacognitive probe data, formally mapped to PAI and MMPI-3 scales. Second, a tiered classification system validated against synthetic response policies. Third, evidence that valid-profile models produce item-sensitive confidence while invalid-profile models do not ($d = 2.17$, $p = .001$). Fourth, chain-of-thought training produces two opposite response distortions with opposite validity profiles. Fifth, the WITHDRAW+BET contradiction as a novel probe-consistency index unique to dual-probe designs. Sixth, retrospective and prospective probes yield inconsistent validity profiles in 12 of 20 models. Seventh, 161 of 486 items significantly discriminate valid from invalid profiles, enabling a short-form screener.

\section{Method}

\subsection{Data}

We analyse 120 CSVs from the Classical Minds metacognitive monitoring battery \citep{cacioli2026c}. Twenty frontier LLMs from six provider families were evaluated: Anthropic (Claude Opus 4.6, Sonnet 4.6, Haiku 4.5), Google-Gemini (Gemini 3 Flash, 3.1 Pro, 2.5 Flash, 2.5 Pro), Google-Gemma (Gemma 1B, 12B, 27B), Qwen (80B Think, 80B Instruct, 235B, Coder 480B), DeepSeek (R1, V3.2), OpenAI (GPT-5.4, 5.4 mini, 5.4 nano), and Zhipu (GLM-5). The battery comprises 524 items across six cognitive domains: learning (T1, 98 items), metacognition (T2, 90 items), social cognition (T3, 116 items), attention (T4, 60 items), executive function (T5, 88 items), and prospective regulation (T6, 72 items). Total evaluations: 10,480.

After every response on T1--T5, two probes were administered: KEEP or WITHDRAW this answer, and BET or NO BET that it is correct. On T6, a prospective path choice (TEXTSC{answer directly}, \textsc{request hint}, or \textsc{decline}) was recorded before the response, followed by retrospective probes.

\subsection{Validity index operationalisation}

Six validity indices were defined by cross-mapping from the PAI and MMPI-3 validity frameworks (Table~\ref{tab:crossmap}).

\begin{table}[ht]
\centering
\small
\caption{Cross-mapping of clinical validity scales to LLM indices.}
\label{tab:crossmap}
\begin{tabularx}{\textwidth}{lllX}
\toprule
\textbf{Index} & \textbf{PAI} & \textbf{MMPI-3} & \textbf{Operationalisation} \\
\midrule
L & PIM & L & $P(\text{KEEP} \mid \text{incorrect})$: maintaining confidence despite errors \\
K & DEF & K & $P(\text{BET} \mid \text{incorrect})$: strong unjustified confidence \\
F & NIM & F & $P(\text{WD} \mid \text{item norm-KEEP} \geq .85)$: withdrawing consensus items \\
Fp & MAL & Fp & $P(\text{WD} \mid \text{correct})$: withdrawing correct answers \\
RBS & RDF & RBS & $P(\text{WD}|\text{correct}) - P(\text{WD}|\text{incorrect})$: inverted monitoring \\
TRIN & --- & TRIN & $\max(n_\text{KEEP}, n_\text{WD}) / n_\text{total}$: fixed responding \\
\bottomrule
\end{tabularx}
\end{table}

For under-reporting, L is defined as $P(\text{KEEP} \mid \text{incorrect})$, measuring maintenance of confidence despite errors. This maps to the PAI's Positive Impression Management (PIM) and the MMPI-3's Uncommon Virtues (L). K is defined as $P(\text{BET} \mid \text{incorrect})$, measuring strong unjustified confidence.

For over-reporting, F is defined as $P(\text{WITHDRAW} \mid \text{item norm-KEEP} \geq .85)$, measuring withdrawal of items that the consensus of models endorses. This assumes that the consensus is informative, a ``wisdom of the crowd'' assumption that will strengthen as the derivation sample grows. Fp is defined as $P(\text{WITHDRAW} \mid \text{correct})$, measuring withdrawal of correct answers.

For response bias, RBS is defined as $P(\text{WD}|\text{correct}) - P(\text{WD}|\text{incorrect})$. A positive value indicates inverted monitoring: withdrawing correct answers more than incorrect ones.

For fixed responding, TRIN is defined as $\max(n_\text{KEEP}, n_\text{WD}) / n_\text{total}$. This operationalisation captures response dominance rather than the alternation inconsistency measured by the MMPI-3's TRIN. It should be interpreted as a proxy for fixed responding. Direction does not matter for validity because both fixed KEEP and fixed WITHDRAW eliminate item-level information.

Additional indices include KEEP-BET concordance (item-level correlation between KEEP and BET decisions), contradiction rate $P(\text{WITHDRAW} \wedge \text{BET})$, and ICN (number of phenotype switches across domains).

L and the withdraw delta ($\Delta w$) share algebraic structure. $L = 1 - P(\text{WD}|\text{incorrect})$. The withdraw delta contains $P(\text{WD}|\text{incorrect})$ as a component. This is not incidental. It is the point. Blanket confidence (high L) mechanically suppresses the monitoring signal. The validity indices detect this suppression. The algebraic relationship is acknowledged throughout and addressed with a non-circular empirical test in Section~\ref{sec:itemsens}.

\subsection{Derivation sample}

This is a derivation study, not a normative validation. The 20-model sample provides initial reference values that require replication on independent model sets.

Item-level reference values were computed for each of the 524 items: $P(\text{KEEP})$ and $P(\text{BET})$ across all 20 models. These define the consensus-endorsed items used in the F index (items where $\geq 85\%$ of models KEEP). Model-level reference values were computed as the 20-model distribution on each index, with mean and SD. Family-level comparison groups were used for exploratory within-family analysis, not for threshold setting.

\subsection{Tiered classification}

Tier 1 (construct-level) uses theory-driven thresholds indicating invalidity regardless of sample distribution. These thresholds are validated against synthetic response policies (Section~\ref{sec:synthetic}). $\text{RBS} > 0$ indicates inverted monitoring. $L \geq 0.95$ indicates near-total failure to detect own errors. $F \geq 0.50$ indicates withdrawal of a majority of consensus-endorsed items. $\text{Fp} \geq 0.50$ indicates withdrawal of a majority of correct answers.

Tier 2 (sample-referenced) is computed after removing Tier 1 invalids. Models are flagged at $M + 1.5\,SD$ (elevated) and $M + 2.0\,SD$ (marked) on the remaining distribution. These thresholds are sample-specific and will shift with larger model sets. The thresholds sit at a stability plateau: sweeping L from 0.93 to 0.97 and varying F/Fp from 0.40 to 0.60 produces the same four Tier 1 classifications across all combinations tested.

Tier 1 means ``do not interpret metacognitive data.'' Tier 2 means ``interpret with caution.''

\subsection{Synthetic policy validation}
\label{sec:synthetic}

To confirm that the tier system correctly separates informative from uninformative response policies, eight synthetic profiles were generated on the same 524-item structure. For each synthetic policy, item-level correctness was drawn from the 20-model mean accuracy per item, preserving the real difficulty distribution. KEEP and BET decisions were generated according to the policy rule independently per item. One thousand bootstrap iterations were run for stochastic policies to produce distributional estimates.

Always KEEP+BET (blanket endorsement) was correctly flagged as invalid ($L \geq .95$). Always WITHDRAW+NO BET (blanket withdrawal) was correctly flagged as invalid ($\text{Fp} \geq .50$). Random 50/50 was correctly flagged as invalid ($\text{RBS}+$, $\text{Fp} \geq .50$). Random 80\% KEEP was correctly passed as valid. Perfect monitor (KEEP correct, WITHDRAW incorrect) was correctly passed. Noisy monitor (80\% KEEP-on-correct, 60\% WITHDRAW-on-incorrect) was correctly passed. Inverted monitor and R1-like (inverted KEEP + always BET) were both correctly flagged.

Bootstrap distributions confirm that random responding produces $\Delta w$ centred at zero ($M = .002$, $SD = .070$) while noisy monitoring produces substantial positive $\Delta w$ ($M = .403$, $SD = .066$), with non-overlapping 95\% confidence intervals.

\begin{figure}[ht]
\centering
\includegraphics[width=\textwidth]{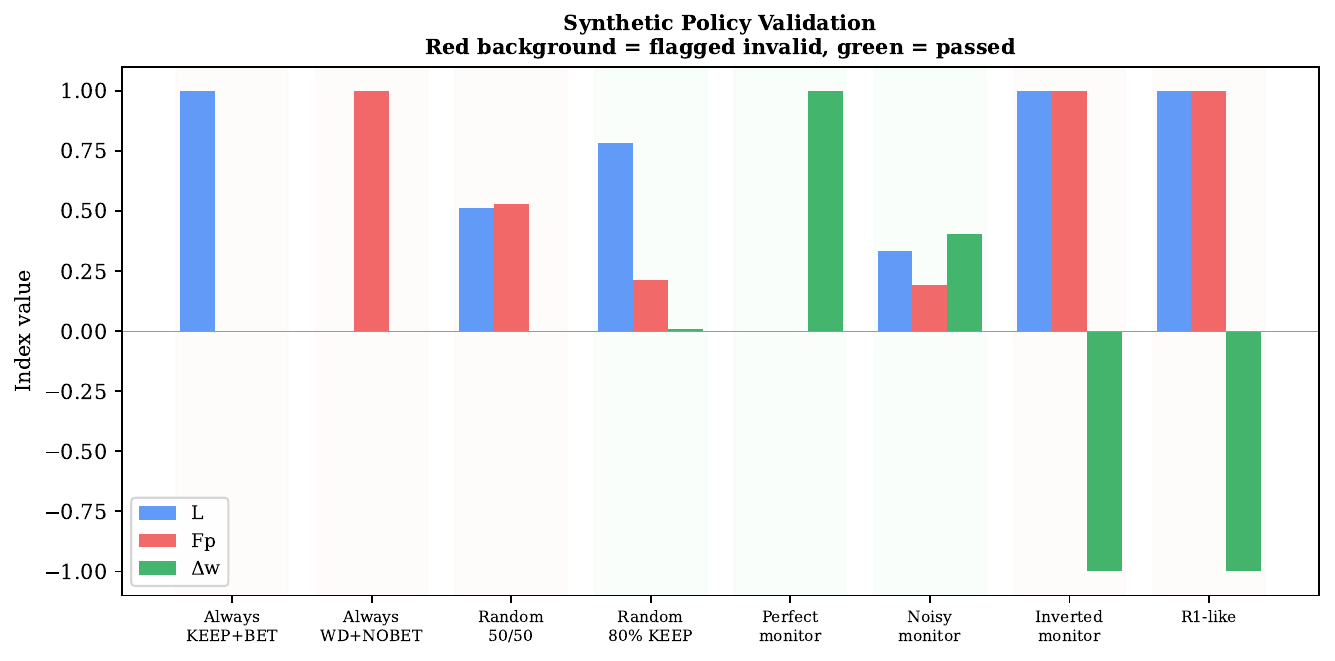}
\caption{Synthetic policy validation. L, Fp, and $\Delta w$ for eight synthetic response policies. Red background indicates policies flagged as Tier 1 invalid. Green indicates policies correctly passed as valid. All uninformative policies are flagged; all informative policies are passed.}
\label{fig:synthetic}
\end{figure}

\subsection{Item-sensitivity analysis}
\label{sec:itemsens}

The central empirical test is whether valid-profile models' KEEP decisions carry item-level information. For each model, the point-biserial correlation between the item-level KEEP decision (0/1) and item-level correctness (0/1) was computed across all 524 items. This metric, $r(\text{KEEP}, \text{correct})$, is not algebraically entangled with any validity index and provides a direct measure of whether the confidence signal tracks performance at the item level.

Valid-profile and invalid-profile models were compared using an independent-samples $t$-test and Cohen's $d$, with bootstrap confidence intervals and leave-one-out sensitivity analysis.

\subsection{Psychometric validation}

Internal consistency was assessed with Cronbach's $\alpha$ for each index across the six tracks. Split-half reliability used odd/even items within tracks, Spearman-Brown corrected. Convergent validity was tested between L and K, between F and Fp, and between $\Delta w$ and BET delta. Discriminant validity was tested between L and F, and between L and accuracy. Factor structure was assessed with PCA on the six indices. Effect sizes used Cohen's $d$ between valid and invalid groups.

\section{Results}

\subsection{Psychometric properties}

The validity indices show strong internal consistency. L across the six tracks yields $\alpha = .921$. K yields $\alpha = .953$. Split-half reliability ranges from $r = .914$ to $.979$ across tracks (Spearman-Brown corrected: $.955$ to $.989$).

Convergent validity is strong within the expected blocs. L and K correlate at $r = .819$ ($p < .0001$). F and Fp correlate at $r = .987$ ($p < .0001$). The withdraw delta and BET delta correlate at $r = .881$ ($p < .0001$).

Discriminant validity is appropriate. L and F correlate at $r = -.103$ (n.s.). L and accuracy correlate at $r = .360$ (n.s.). The two dimensions are independent, consistent with the clinical finding that under-reporting and over-reporting represent separable response styles.

\subsection{Factor structure}

PCA on the six indices yields two components explaining 94.6\% of variance. PC1 (50.2\% variance) loads on Fp ($+.56$), F ($+.55$), and TRIN ($-.44$). This is the over-reporting dimension. PC2 (44.4\% variance) loads on K ($-.55$), L ($-.55$), and RBS ($-.48$). This is the under-reporting dimension. The third component accounts for 4.1\% of variance.

Two latent dimensions capture essentially all systematic variance. This mirrors the clinical assessment finding that validity scales cluster into under-reporting and over-reporting blocs with relative independence between them. The clean two-factor structure validates the transfer of the PAI/MMPI-3 organisational framework to LLM probe data.

\begin{figure}[ht]
\centering
\includegraphics[width=0.65\textwidth]{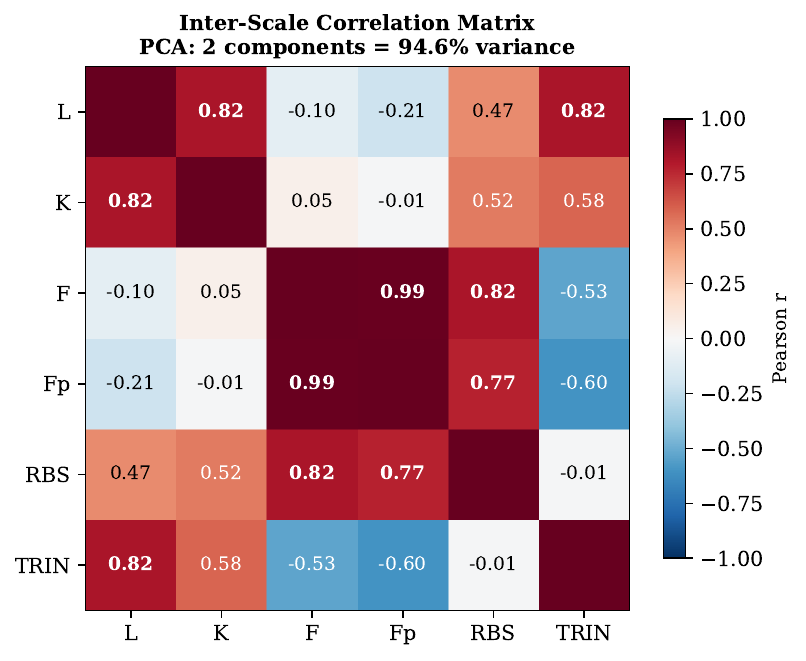}
\caption{Inter-scale correlation matrix. L and K cluster as the under-reporting bloc ($r = .82$). F and Fp cluster as the over-reporting bloc ($r = .99$). L and F are independent ($r = -.10$). PCA confirms two components accounting for 94.6\% of variance.}
\label{fig:corr}
\end{figure}

\subsection{Synthetic policy validation}

All eight synthetic policies were correctly classified by the tier system. Uninformative policies were flagged. Informative policies were passed. The mildly biased random 80\% KEEP policy was correctly passed: its $\Delta w$ centres at zero, indicating no monitoring signal despite the endorsement bias. The framework detects extreme response sets that suppress item-level variance. Mildly biased policies may pass Tier 1 because they do not eliminate variance, but they are identifiable through the item-sensitivity test (Section~\ref{sec:res_itemsens}).

\subsection{Tiered classification}

Tier 1 invalid ($n = 4$): DeepSeek-R1 (RBS+, $F \geq .50$, $\text{Fp} \geq .50$; $z$-scores on over-reporting scales: $+3.7$ to $+4.0$), Gemini 3.1 Pro (RBS+, $L \geq .95$), Qwen 80B Think ($L \geq .95$; $L = .974$), and Gemma 1B (RBS+; $\Delta w = -.028$).

Tier 2 elevated ($n = 2$): GPT-5.4 nano (F and Fp elevated, but positive $\Delta w = +.145$) and Gemma 12B (Fp borderline elevated, with $\Delta w = +.248$).

Valid ($n = 14$): the remaining models.

\begin{figure}[ht]
\centering
\includegraphics[width=0.85\textwidth]{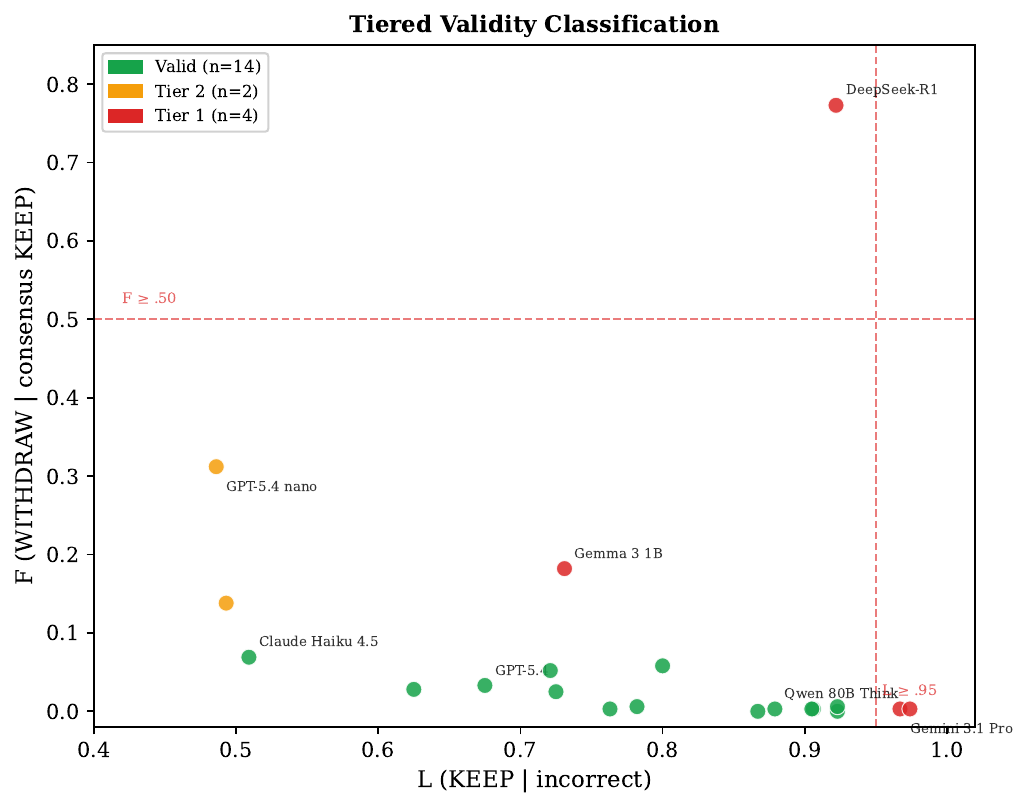}
\caption{Tiered validity classification. L (KEEP on errors) versus F (WITHDRAW on consensus items) for all 20 models. Dashed lines show Tier 1 thresholds ($L \geq .95$, $F \geq .50$). DeepSeek-R1 occupies the extreme upper-right quadrant. Qwen 80B Think sits against the right threshold. Valid models cluster in the lower-left.}
\label{fig:tiered}
\end{figure}

\subsection{Item sensitivity}
\label{sec:res_itemsens}

This is the central empirical test. Valid-profile models ($n = 16$, including Tier 2 elevated) show a mean $r(\text{KEEP}, \text{correct}) = .180$ ($SD = .058$). Fourteen of sixteen produce individually significant positive correlations. Their KEEP decisions systematically track whether they got the item right.

Invalid-profile models ($n = 4$) show a mean $r(\text{KEEP}, \text{correct}) = -.196$ ($SD = .403$). None produce individually significant positive correlations. DeepSeek-R1 shows $r = -.798$, a massive inversion. Gemini 3.1 Pro shows $r = -.001$. Gemma 1B shows $r = -.031$. Qwen 80B Think shows $r = .047$.

The group comparison yields Cohen's $d = 2.17$, $t(18) = 3.89$, $p = .001$. The bootstrap 95\% confidence interval on $d$ is $[1.95, 4.72]$. The large effect size reflects near-complete separation between models that preserve item-level variance and those that collapse to response sets, rather than fine-grained differences within a continuous distribution.

The result is robust. Without R1, $d$ increases to $3.10$ ($p = .0001$) because the invalid group's SD shrinks. Leave-one-out analysis confirms significance ($p < .05$) for every single model removal. No individual model drives the result.

Invalid-profile models produce confidence signals that carry little or no item-level information about correctness. This is the empirical justification for screening.

\begin{figure}[ht]
\centering
\includegraphics[width=0.85\textwidth]{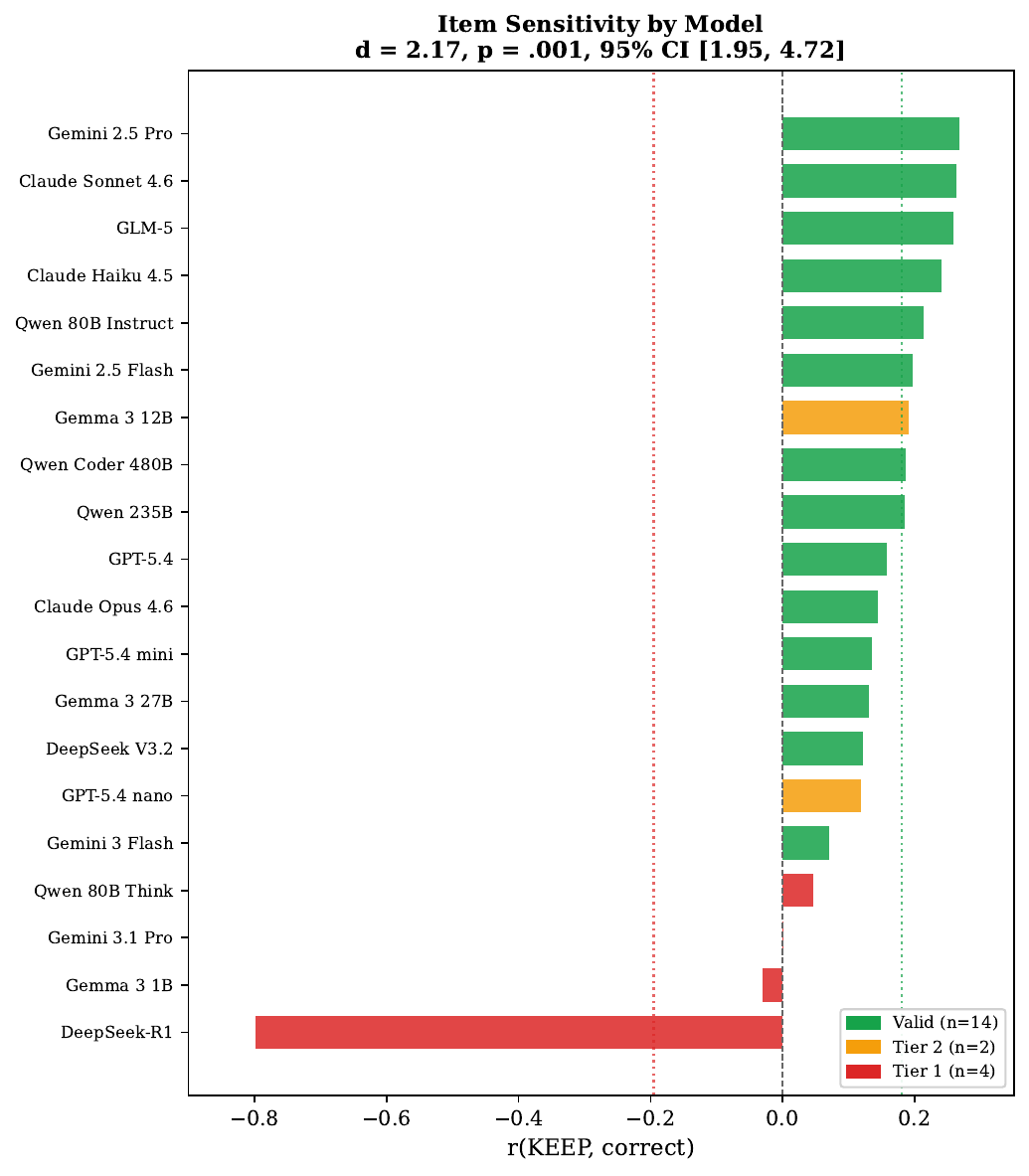}
\caption{Item sensitivity: $r(\text{KEEP}, \text{correct})$ by model. Valid models (green) show positive item sensitivity. Tier 1 invalid models (red) show zero or negative. The group difference is $d = 2.17$ ($p = .001$), robust to leave-one-out removal of any single model. Bootstrap 95\% CI on $d$: $[1.95, 4.72]$.}
\label{fig:itemsens}
\end{figure}

\subsection{Incremental prediction}

L and $\Delta w$ share algebraic components. $L = 1 - P(\text{WD}|\text{incorrect})$. The withdraw delta contains $P(\text{WD}|\text{incorrect})$. This is acknowledged.

Accuracy alone predicts $\Delta w$ at $R^2 = .032$ (n.s.). Accuracy plus L predicts at $R^2 = .357$ ($\Delta R^2 = .325$, $F(1,17) = 8.59$, $p = .009$). Higher L mechanically suppresses $\Delta w$. L does not predict $\Delta w$ incidentally. It predicts it because blanket confidence eliminates the differential withdrawal that constitutes monitoring.

As a non-circular alternative, item-level $r(\text{KEEP}, \text{correct})$ was used as the dependent variable. Accuracy alone: $R^2 = .071$. Accuracy plus L: $R^2 = .237$ ($\Delta R^2 = .166$, $F(1,17) = 3.71$, $p = .071$). The direction is consistent but the effect is underpowered at $n = 20$. The item-sensitivity group comparison provides the stronger, non-circular test.

\subsection{The WITHDRAW+BET contradiction}

DeepSeek-R1 produces WITHDRAW+BET on 32.6\% of all items. On correct items: 37.1\%. No other model exceeds 0.6\%.

R1 simultaneously says ``I withdraw this answer'' and ``I bet it is correct.'' The monitoring signal (WITHDRAW) and the confidence signal (BET) are decoupled. This is a novel probe-consistency index enabled by the dual-probe design. Structurally, the model produces a correct answer and then retracts it while simultaneously betting on it. The WITHDRAW and BET signals are decoupled.

The contingency tables are stark. R1: 77 KEEP+BET, 0 KEEP+NO BET, 171 WITHDRAW+BET, 0 WITHDRAW+NO BET. Every response includes BET regardless of the KEEP/WITHDRAW decision. Claude Haiku 4.5 (highest monitoring): 355 KEEP+BET, 0 KEEP+NO BET, 0 WITHDRAW+BET, 0 WITHDRAW+NO BET. When Haiku withdraws, it does not bet. The probes are fully coupled.

\begin{figure}[ht]
\centering
\includegraphics[width=\textwidth]{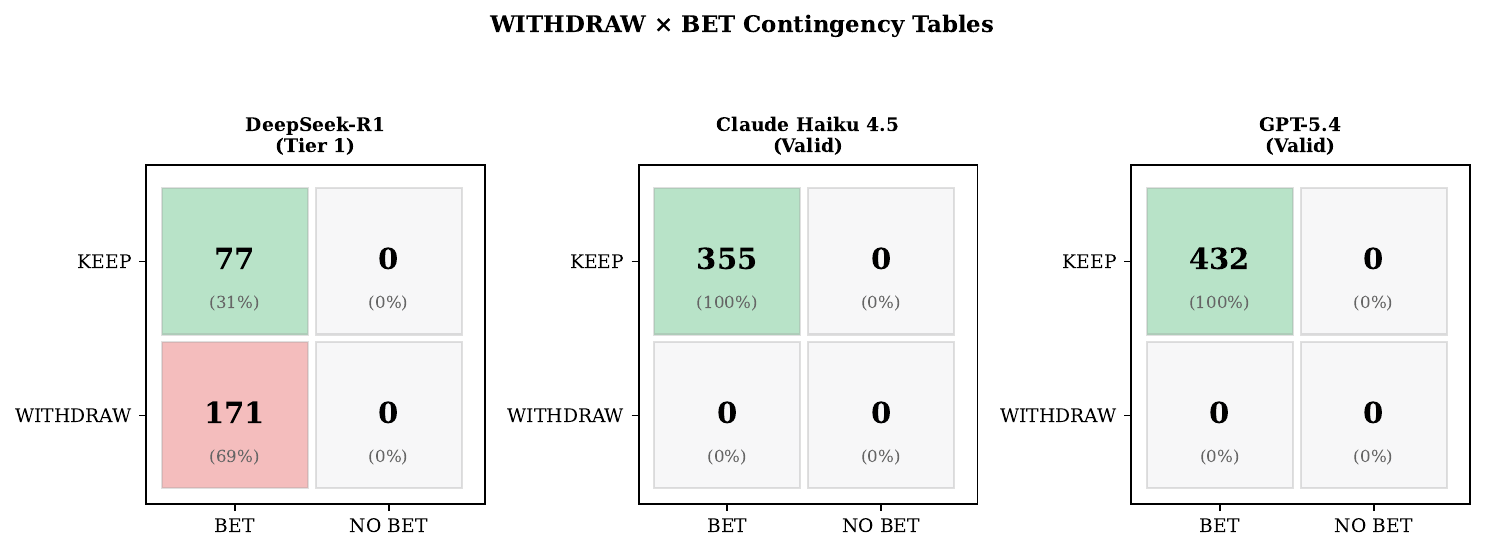}
\caption{WITHDRAW $\times$ BET contingency tables for DeepSeek-R1, Claude Haiku 4.5, and GPT-5.4. R1 shows 171 WITHDRAW+BET responses (red), indicating fully decoupled probes. In all other models, WITHDRAW and BET are coupled: models that withdraw do not bet.}
\label{fig:contingency}
\end{figure}

\subsection{Reasoning mode produces opposite distortions}

Two paired comparisons isolate the effect of chain-of-thought training within the same architecture.

DeepSeek R1 versus V3.2. Same architecture, R1 adds chain-of-thought. WITHDRAW on correct items shifts from 8.6\% to 94.6\%, a delta of $+86$ percentage points. K shifts from .317 to .987 ($+67$ pp). The contradiction rate on correct items shifts from 0.0\% to 37.1\%. The withdraw delta shifts from $+.114$ to $-.868$, a swing of nearly one full unit.

Qwen 80B Think versus Instruct. Same base model, Think adds reasoning mode. L shifts from .782 to .974 ($+.193$). The withdraw delta shifts from $+.171$ to $+.017$ ($-.154$). Monitoring collapses from genuine sensitivity to near-zero.

Chain-of-thought training does not uniformly improve or degrade validity. It produces opposite distortions depending on implementation.

\begin{figure}[ht]
\centering
\includegraphics[width=\textwidth]{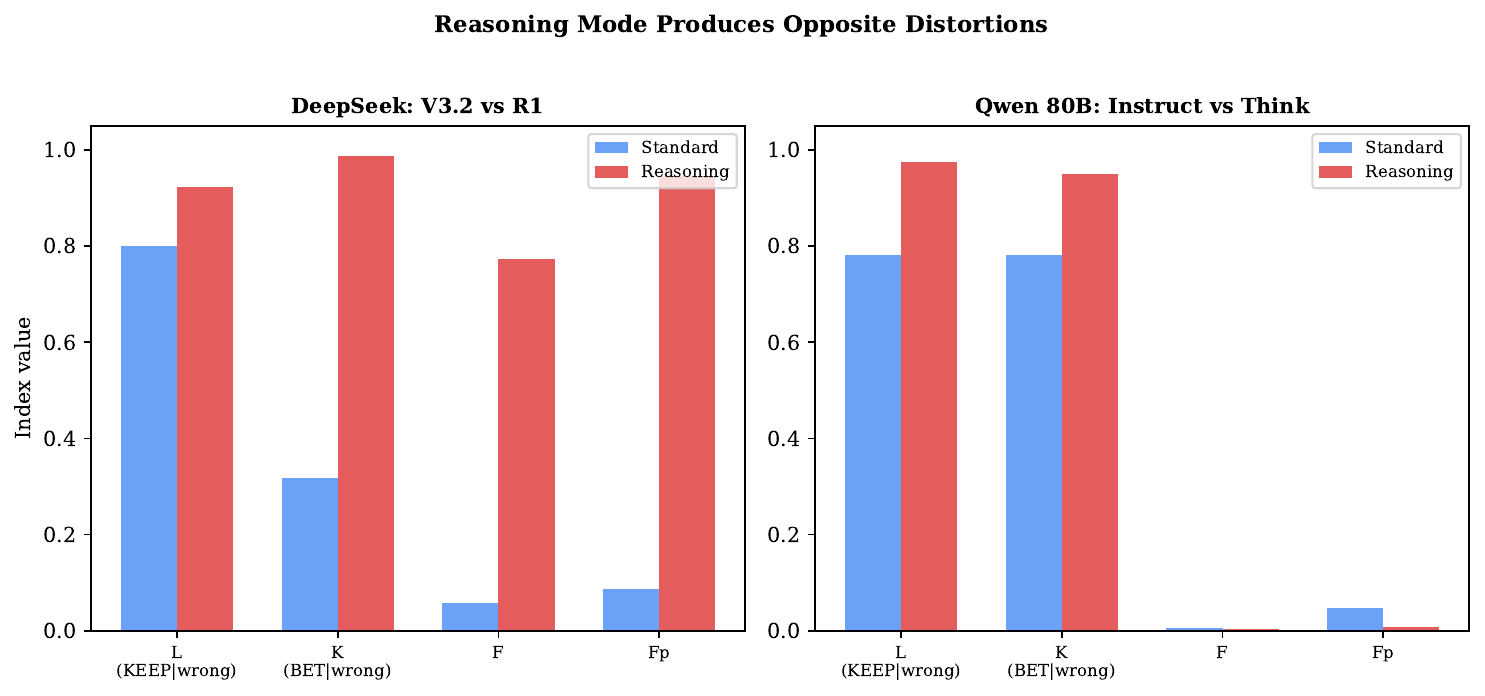}
\caption{Reasoning mode produces opposite distortions. Same architecture, different training. R1 develops extreme over-reporting (high F, Fp) while V3.2 is moderate. Qwen Think develops extreme under-reporting (high L, K) while Instruct shows genuine monitoring.}
\label{fig:reasoning}
\end{figure}

\subsection{Retrospective versus prospective validity}

Twelve of 20 models show inconsistent validity profiles across retrospective (T1--T5) and prospective (T6) probe formats. Multiple models show $L_\text{retro} \approx .40$--.72 but $L_\text{prosp} = 1.00$. They monitor retrospectively but not prospectively.

T6 items produce almost no validity discrimination. Of 40 prospective items, 1 significantly discriminates valid from invalid profiles. Of 88 executive function items (T5), 41 discriminate.

Validity is not a stable property of the model but of the interaction between model, probe format, and response policy. The T6 result may reflect the difference between evaluating a completed response (retrospective) and committing to a strategy before responding (prospective). The prospective format may constrain response variance in ways that prevent validity discrimination.

\subsection{Family-level comparison groups}

These are exploratory observations from small within-family samples ($n = 2$ to $4$), not claims about stable family traits.

Google-Gemini ($n = 4$) shows a uniform under-reporting profile. L ranges from .906 to .967 ($M = .930$, $SD = .026$). K ranges from .872 to .906. F and Fp are near zero in all four models. This consistency suggests a shared training effect, structurally analogous to coaching effects in forensic assessment \citep{sellbom2008,rogers2003}.

OpenAI ($n = 3$) shows a gradient with scale. L decreases from .721 (mini) to .675 (5.4). $\Delta w$ increases from $+.145$ (nano) to $+.203$ (5.4). Scaling improves both accuracy and monitoring.

Anthropic ($n = 3$) shows a gradient from Opus (defensively confident: $L = .879$, $\Delta w = +.099$) to Haiku (genuine monitor: $L = .509$, $\Delta w = +.316$). The least accurate Anthropic model has the best monitoring.

\subsection{Item-level validity discriminators}

161 of 486 items (33\%) significantly discriminate valid from invalid profiles at $p < .05$ (point-biserial correlation, uncorrected). T5 (executive function): 41 of 88 items. T2 (metacognition): 34 of 86. T4 (attention): 31 of 60. T3 (social cognition): 33 of 114. T1 (learning): 21 of 98. T6 (prospective): 1 of 40. These are exploratory, post-hoc, and require cross-validation on new models.

\section{Discussion}

\subsection{The structural parallel holds}

The indices show the same convergent and discriminant structure as their clinical counterparts. L and K cluster as an under-reporting bloc. F and Fp cluster as an over-reporting bloc. L and F are independent. The factor structure confirms two latent dimensions accounting for 94.6\% of variance, mirroring the clinical organisation of validity scales.

This structure emerges not because LLMs have personalities but because the measurement problem is structurally identical. Detecting when binary self-report data is dominated by a response set rather than item content is the same statistical problem regardless of whether the respondent is a human or a language model.

\subsection{Policy artefacts versus informative signals}

The item-sensitivity analysis provides the empirical answer to the policy objection. We do not claim that valid-profile models introspect. We show that their confidence outputs carry item-level information that invalid-profile models' outputs do not. Whether the information source is introspection, learned statistical regularities, or something else is irrelevant to whether the signal is usable.

The validity framework is therefore not a screen for genuine metacognition. It is a screen for informative confidence. This is a weaker but more defensible claim, and it is the claim that matters for deployment.

\subsection{Response distortion profiles}

Six distinct response patterns emerge from the data. These are described using structural labels. The clinical analogues describe the geometry of the response pattern, not psychological traits of the respondent. Blanket confidence (Qwen Think, Gemini 2.5 Flash/Pro): high L, high K, near-zero F. Inverted monitoring with contradiction (DeepSeek-R1): withdraws correct answers, bets on incorrect ones. Uninformative responding (Gemini 3.1 Pro, Gemma 1B): near-zero $\Delta w$. Uniform high-confidence (Google-Gemini family): uniform high L and K across four models. Genuine monitoring (Haiku, Sonnet, GPT-5.4, Qwen Coder/Instruct/235B): positive $\Delta w$, moderate L, item-sensitive confidence. Elevated withdrawal (GPT-5.4 nano): high F and Fp, but genuine monitoring signal.

\subsection{RLHF as a response set inducer}

RLHF optimises for human preference, which rewards confidence. This is structurally analogous to coaching effects in forensic assessment \citep{sellbom2008,rogers2003}. The Gemini family's uniform defensive profile across four models may be an RLHF artefact.

Chain-of-thought training is a different kind of coaching. It teaches the model to deliberate, but the deliberation can become structurally distorted. In R1, it produces excessive withdrawal despite high accuracy. In Think, it produces excessive endorsement despite errors. These are opposite distortions arising from the same training category.

\subsection{Implications for deployment}

For selective prediction, an L-elevated model produces no abstention signal. Threshold-based abstention systems built on such models will either abstain on nothing or abstain randomly.

For safety, a model that looks 94\% accurate but produces uninformative confidence is more dangerous than one that is 88\% accurate with genuine monitoring. The high-accuracy model provides false assurance that confidence-based safeguards are working.

For benchmarking, validity screening should precede interpretation of any metacognitive metric. The item-level discriminators identified here provide a starting point for a short-form validity screener embeddable in future benchmarks. A portable screening protocol extracting a minimal index set from these results is presented in the companion paper (Cacioli, 2026e).

\subsection{Limitations}

This is a derivation study. $N = 20$ models. All thresholds are sample-specific and require replication. Family-level comparisons ($n = 2$--$4$) are exploratory. Single administration with no test-retest reliability. Binary probes only. Algebraic entanglement between L and $\Delta w$ is acknowledged; the item-sensitivity analysis provides the non-circular test but is underpowered at $n = 20$. No external criterion for threshold calibration in the present paper; concurrent criterion validation against selective prediction is reported in Cacioli (2026f). Causal mechanism unknown. Item-level discriminators are post-hoc and require cross-validation.

\section{Conclusion}

The clinical assessment literature has spent 50 years solving the problem of response validity in self-report data. LLM evaluation faces the same structural problem and has not borrowed the solution. We show that the solution transfers: validity indices, tiered classification, synthetic policy validation, and the fundamental principle that you screen before you interpret. The central finding is not that some models have invalid personalities. It is that some models produce confidence signals that carry little or no item-level information about correctness, and that this can be detected before substantive interpretation begins.

\section*{Open science}

All 524 items, 120 CSVs, and analysis code are publicly available at \url{https://github.com/synthiumjp/validity-scaling-llm}. The battery and its companion paper are described in \citet{cacioli2026c}. The portable screening protocol is specified in Cacioli (2026e). Concurrent criterion validation against selective prediction is reported in Cacioli (2026f).

\section*{Generative AI disclosure}

Claude (Anthropic) was used for analysis pipeline design, code generation, and manuscript preparation. All scientific decisions were made by the author.

\bibliographystyle{apalike}
\bibliography{references}

\end{document}